\def\year{2020}\relax
\documentclass[letterpaper]{article} 
\usepackage{aaai20}  
\usepackage{times}  
\usepackage{helvet} 
\usepackage{courier}  
\usepackage[hyphens]{url}  
\usepackage{graphicx} 
\usepackage{graphicx}  
\usepackage{listings}

\usepackage{amsfonts,amsmath}
\usepackage{multirow}
\usepackage{xcolor}

\usepackage{algorithm}
\usepackage{algorithmic}
\usepackage[font=small,format=hang,parskip=1pt]{caption}

\newcommand{\citet}[1]{\citeauthor{#1} \shortcite{#1}}
\newcommand{\citep}{\cite}

\usepackage[algo2e,ruled,linesnumbered,vlined]{algorithm2e}

\newcommand{\argmin}{\operatornamewithlimits{argmin}}

\newcommand{\Diff}{\emph{Diff}}

\newtheorem{definition}{{\bf Definition}}

\newtheorem{example}{\bf Example}

\newcommand{\bitemize}{\begin{list}{$\bullet$}{\topsep=1pt \parsep=0pt \itemsep=1pt \leftmargin=1em }} 
\newcommand{\eitemize}{\end{list}}
\newcommand{\beitemize}{\begin{list}{$\bullet$}{\topsep=1.5pt \parsep=0pt \itemsep=1pt \leftmargin=1em }} 
\newcommand{\enitemize}{\end{list}}

\def\credulous{{\models^c_L}}
\def\skeptical{{\models^s_L}} 

\title{On the Relationship Between KR Approaches for Explainable Planning}
\author{
  Stylianos Loukas Vasileiou,\textsuperscript{\rm 1}
        William Yeoh \textsuperscript{\rm 1} and 
        Tran Cao Son \textsuperscript{\rm 2} \\
        \textsuperscript{\rm 1} Computer Science \& Engineering, Washington University in St. Louis \\
         \textsuperscript{\rm 2} Computer  Science, New Mexico State University \\
         v.stylianos@wustl.edu, wyeoh@wustl.edu, stran@nmsu.edu}

\begin{document}

\maketitle
\sloppy
\allowdisplaybreaks

\begin{abstract}
\small
In this paper, we build upon notions from \emph{knowledge representation and reasoning} (KR) to expand a preliminary logic-based framework that characterizes the model reconciliation problem for explainable planning. We also provide a detailed exposition on the relationship between similar KR techniques, such as abductive explanations and belief change, and their applicability to explainable planning.

\end{abstract}

%
%

\section{Introduction}
\label{sec:intro}
As there is a substantial need for transparency and trust between intelligent systems and humans, \emph{Explainable AI Planning} (XAIP) has recently gained a lot of attention due to its potential adoption in real-world applications. Within this context, a popular theme that has recently emerged is called \emph{model reconciliation}~\cite{chakraborti2017plan}. Researchers in this area have looked at how an agent can explain its decisions to a human user who might have a different understanding of the same planning problem. These explanations bring the model of the human user closer to the agent's model by transferring a minimum number of updates from the agent's model to the human's model. However, a common thread across most of these works is that they, not surprisingly, employ mostly automated planning approaches. 

In this work, we tackle the \emph{model reconciliation problem} (MRP) from the perspective of \emph{knowledge representation and reasoning} (KR), where we lay the theoretical foundations that extend a logic-based framework proposed by \citet{vas} and argue that it can effectively model the MRP. As our framework builds upon various KR techniques, we further give a detailed exposition on the relationship between such techniques and their applicability to XAIP as well as provide two examples that highlight the differences with our proposed framework.

\section{Background} 

\medskip \noindent \textbf{Logic:}
A \emph{logic} ${L}$ is a tuple $\langle KB_L, BS_L, ACC_L \rangle$, where  
 ${KB_L}$ is the set of well-formed knowledge bases (or theories)
 of ${L}$ -- each being  a set of formulae. 
 ${BS_L}$ is the set of possible belief sets; each element of ${BS_L}$ is a set of syntactic 
 elements representing the beliefs $L$ may adopt.  
 $ACC_L: KB_L \rightarrow 2^{{BS_L}}$ describes the \emph{``semantics''} 
 of $L$ by assigning to each element of ${KB_L}$ a set of acceptable sets of beliefs. For 
 each $KB \in KB_L$ and $B \in ACC_L(KB)$, we say that $B$ is a \emph{model} of $KB$. 
 A logic is monotonic if $KB \subseteq KB'$ implies $ACC_L(KB') \subseteq ACC_L(KB)$. 
 
 
 \begin{definition}
 [Skeptical Entailment]
 A formula $\varphi$ in the logic $L$ is \emph{skeptically entailed} by $KB$, 
 denoted by $KB \skeptical  \varphi$,  
 if $ACC_L(KB) \ne \emptyset$ and $\varphi \in B$ for \emph{every} $B \in ACC_L(KB)$.
\end{definition}

 
 \begin{definition}
 [Credulous Entailment]  
 A formula $\varphi$ in the logic $L$ is \emph{credulously entailed} by $KB$, 
 denoted by $KB \credulous  \varphi$,  
 if $ACC_L(KB) \ne \emptyset$ and $\varphi \in B$ for \emph{some} $B \in ACC_L(KB)$
 \end{definition} 
 

\begin{definition}
[Consistent Knowledge Base]
A $KB$ is \emph{consistent} iff $ACC_L(KB) \neq \emptyset$ or, equivalently, iff $KB$ does not skeptically entail false. 
\end{definition}
 
Throughout this paper, we consider a finitary propositional language $L$ and represent a knowledge base by a propositional formula $KB$. For our later use, we will assume that a negation operator $\neg$ over formulas exists.   
Additionally, $\varphi$ and $\neg \varphi$ are contradictory with each other in the sense that, 
for any $KB$ and $B \in ACC_L(KB)$, if $\varphi \in B$, then $\neg \varphi \not\in B$; 
and if $\neg \varphi \in B$, then $\varphi \not\in B$. Therefore, if $\{\varphi, \neg \varphi\} \subseteq KB$, then $KB$ is inconsistent, i.e.,~$ACC_L(KB) = \emptyset$. 
$\epsilon \subseteq KB$ is called a \emph{sub-theory} of $KB$. 
A theory $KB$ \emph{subsumes} a theory $KB'$, denoted by $KB \lhd KB'$, if 
$ACC_L(KB) \subset ACC_L(KB')$.

\medskip \noindent \textbf{Classical Planning:}A classical planning problem, typically represented in PDDL~\cite{ghallab1998pddl}, is a tuple $ \Pi = \langle D, I, G\rangle$, which consists of the domain $D = \langle F, A \rangle$ -- where $F$ is a finite set of fluents representing the world states ($s\in F$) and $A$ a set of actions -- and the initial and goal states $I, G \subseteq F$. An action $a$ is a tuple $\langle pre_{a}, \textit{eff}_{a}\rangle$, where $pre_{a}$ are the preconditions of $a$ -- conditions that must hold for the action to be applied; and $\textit{eff}_{a} = \langle \textit{eff}_{a}^{+}, \textit{eff}_{a}^{-} \rangle$ are the addition ($ \textit{eff}_{a}^{+}$) and deletion ($ \textit{eff}_{a}^{-}$) effects of $a$ -- conditions that must hold after the action is applied. More formally, using $\delta_{\Pi}: 2^F \times A \rightarrow 2^F$ to denote the transition function of problem $\Pi$, if $s \not \models pre_{a}$, then $\delta _{\Pi}(s, a) \models \perp$; otherwise, $\delta _{\Pi}(s, a) \models s \cup \textit{eff}^{+}_{a} \setminus \textit{eff}^{-}_{a}$. The solution to a planning problem $\Pi$ is a plan $\pi = \langle a_{1}, \ldots , a_{n} \rangle$ such that $\delta _{\Pi}(I, \pi) \models G$, 
where 
$\delta_\Pi(s, \pi) = \delta_\Pi(\delta_\Pi(s,a_1), \pi')$ with $\pi' = \langle a_2,\ldots, a_n\rangle$.
The cost of a plan $\pi$ is given by $C(\pi, \Pi) =  |\pi|$. 
Finally, the cost-minimal plan $\pi^{*} = \argmin_{\pi \in \{\pi' \mid \delta_{\Pi}(I, \pi') \models G\}}  C(\pi, \Pi)$ is called the optimal plan.

\medskip \noindent \textbf{Classical Planning as Boolean Satisfiability:}
A classical planning problem can be encoded as a SAT problem~\cite{kautz1992planning,kautz1996encoding}. The basic idea is the following: Given a planning problem $P$, find a solution for $P$ of length $n$ by creating a propositional formula that represents the initial state, goal state, and the action dynamics for $n$ time steps. This is referred to as the \emph{bounded planning problem} $(P,n)$, and we define the formula for $(P,n)$ such that: \emph{Any} model of the formula represents a solution to $(P,n)$ and if $(P,n)$ has a solution, then the formula is satisfiable.

We encode $(P,n)$ as a formula $\Phi$ involving one variable for each action $a \in A$ at each timestep $t \in \{ 0, \ldots, n-1\}$ and one variable for each fluent $f \in F$ at each timestep $t \in \{0, \ldots, n \}$. We denote the variable representing action $a$ in timestep $t$ using subscript $a_t$, and similarly for facts. The formula $\Phi$ is constructed such that $\langle a_0, a_1, \ldots , a_{n-1} \rangle$ is a solution for $(P,n)$ if and only if $\Phi$ can be satisfied in a way that makes the fluents $a_0, a_1, \ldots , a_{n-1}$ true. The formula $\Phi$ is a conjunction of the following formulae: 
\bitemize
\item \textbf{Initial state:} Let $F$ and $I$ be the sets of fluents and initial states, respectively, in the planning problem:  
\begin{align}
\bigwedge_{ f \in I } f_0  \wedge \bigwedge_{f \in F \setminus \{ I\}} \neg f_0 
\end{align}

\item \textbf{Goal state:} Let $G$ be the set of goal states:
\begin{align}
\bigwedge_{f \in G } f_n 
\end{align}

\item \textbf{Action scheme:} Formulae enforcing the preconditions and effects of each action $a$ at time step $t$:
\begin{align}
 a_t &\Rightarrow \bigwedge_{f \in pre_{a} }f_t  \\
 a_{t} &\Rightarrow \bigwedge_{f \in \textit{eff}^{+}_{a}} f_{t+1}  \\
 a_{t} &\Rightarrow \bigwedge_{f \in \textit{eff}^{-}_{a}} \neg f_{t+1}
\end{align}

\item \textbf{Explanatory frame axioms:}  Formulae enforcing that facts do not change between subsequent time steps $t$ and $t+1$ unless they are effects of actions that are executed at time step $t$:
\begin{align}
\neg f_t \wedge f_{t+1} &\Rightarrow \bigvee \{ a_t \mid f \in \textit{eff}^{+}_{a} \} \\
f_t \wedge \neg f_{t+1} &\Rightarrow \bigvee \{ a_t \mid f \in \textit{eff}^{-}_{a} \}
\end{align}

\item \textbf{Action exclusion axioms:} Formulae enforcing that only one action can occur at each time step $t$:
\begin{align} 
& \bigwedge_{a \in A}\bigwedge_{a' \in A \mid a \not =  a'} (\lnot a_{t} \vee \lnot a_{t}')
\end{align}
where $A$ is the set of actions in the planning problem. 

\eitemize

Finally, we can \emph{extract} a plan by finding an assignment of truth values that satisfies $\Phi$ (i.e.,~for all time steps $t=0, \ldots, n-1$, there will be exactly one action $a$ such that $a_t = $ \emph{True}). This could be easily done by using a satisfiability algorithm, such as the well-known DPLL algorithm~\cite{DPLL62}.

\medskip \noindent \textbf{Model Reconciliation Problem:}
 An MRP is defined by the tuple $\Psi = \langle \Phi, \pi  \rangle $, where $\Phi = \langle M^R, M^{R}_{H} \rangle$ is a tuple of the agent's model $M^R = \langle D^R, I^R, G^R \rangle$ and the agent's approximation of the human's model $M^{R}_H = \langle D^R_H, I^R_H, G^R_H \rangle$ , and $\pi$ is the optimal plan in $M^R$. A solution to an MRP is an explanation $\epsilon$ such that when it is used to update the human's model $M^{R}_H$ to $\widehat{M}^{R, \epsilon}_H$, the plan $\pi$ is optimal in both the agent's model $M^R$ and the updated human model $\widehat{M}^{R, \epsilon}_H$. The goal is to find a cost-minimal explanation, where the cost of an explanation is defined as the length of the explanation by \citet{chakraborti2017plan}

\section{Logic-based Explanations in Planning} 

We now describe our framework, which generalizes the preliminary framework proposed by \citet{vas}, that solves the model reconciliation problem by computing cost-minimal explanations with respect to two knowledge bases. We formulate the notion of explanation in the following setting, where, for brevity, we use the term $\models_{L}^{x}$ for $x\in \{s,c \}$ to refer to skeptical ($s$) or credulous ($c$) entailment: 
\begin{quote} 
{\bf Explanation Generation Problem:} Given two knowledge bases $KB_a$ and $KB_h$  and a formula $\varphi$ in a logic $L$.
Assume that $KB_a \models_{L}^{x} \varphi$ and $KB_h  \not\models_{L}^{x} \varphi$. 
The goal is to identify an explanation (i.e.,~a set of formulas) $\epsilon \subseteq KB_a$ such that when it is used to \emph{update} $KB_h$ to $\widehat{KB}_h^{\epsilon}$, the updated $\widehat{KB}_h^{\epsilon} \models_{L}^{x} \varphi$. 
\end{quote}

When updating a knowledge base $KB$ with an explanation $\epsilon$, the updated knowledge base $KB \cup \epsilon$ may be inconsistent as there may be contradictory formulas in $KB$ and $\epsilon$. As such, to make $KB$ consistent again, one needs to remove this set of contradictory formulae $\gamma \subseteq KB$ from $KB$. Further, it is vital that the $KB$ must contain all formulae pertaining to the action dynamics of the given planning problem (see~\cite{kautz1996encoding}). 

\begin{definition}
[Knowledge Base Update]
\label{def:update}
Given a knowledge base $KB$ and an explanation $\epsilon$, the updated knowledge base is $\widehat{KB}^{\epsilon} = KB \cup \epsilon \setminus \gamma$, where $\gamma \subseteq KB\setminus \epsilon$ is a set of formulae that must be removed from $KB$ such that the updated $\widehat{KB}^{\epsilon}$ is consistent.\footnote{Intuitively, one should prefer the set of formula $\gamma$ that is removed to be as small as possible, though we chose to not require such a restriction here.}
\end{definition}
We now define the notion of a \emph{support} of a formula w.r.t. a knowledge base before defining the notion of \emph{explanations}.
\begin{definition} 
[Support]
\label{def:support}
Given a knowledge base $KB$ and a formula $\varphi$ in logic $L$, assume that $KB \models_{L}^{x} \varphi$.  
 We say that $\epsilon \subseteq KB$ is an \emph{$x$-support} 
  of $\varphi$  w.r.t. $KB$  if $\epsilon \models_{L}^{x} \varphi$.
 Assume that $\epsilon$ is an $x$-support of $\varphi$ w.r.t. $KB$. 
We say that $\epsilon \subseteq KB$ is a \emph{$\subseteq$-minimal $x$-support} of $\varphi$ if 
 no proper sub-theory of $\epsilon$ is an $x$-support of $\varphi$. 
Furthermore,  
 $\epsilon$ is a \emph{$\lhd$-general $x$-support} of $\varphi$ if 
 there is no support $\epsilon'$ of $\varphi$ w.r.t. $KB$ 
 such that $\epsilon$ subsumes $\epsilon'$. 
 
\end{definition}

\begin{definition}
[Explanation] 
\label{def:m-exp}
Given two knowledge bases $KB_a$ and $KB_h$ and a formula $\varphi$ in logic $L$, assume that $KB_a \models_{L}^{x}  \varphi$ and $KB_h  \not \models_{L}^{x}  \varphi$. An \emph{explanation for $\varphi$ from $KB_a$ for $KB_h$} 
is a support $\epsilon$ w.r.t. $KB_a$ for $\varphi$ such that the updated $\widehat{KB}_h^{\epsilon} \models_{L}^{x} \varphi$, where $\widehat{KB}_h^{\epsilon}$ is updated according to Definition~\ref{def:update}. 
\end{definition}

\subsection{Explanations in Planning}
Even though explanations can be composed for arbitrary queries, in this paper, we identify two important problems: (1)~Explaining the \emph{validity} of a plan to the human user, and (2)~Explaining the \emph{optimality} of a plan to the user. Naturally, the former problem must be solved before the latter problem since the user must accept that the plan is valid before they accept that the plan is optimal. However, it may be the case that only the former problem must be solved, especially when plan optimality is not a major concern to the user. From now until the end of this section, we use $KB_a$ and $KB_h$ to denote the knowledge bases encoding the planning problem of the planning agent and the human user, respectively.

\medskip \noindent \textbf{Plan Validity:}
Assume $\pi$ is a valid plan with respect to $KB_a$ but not $KB_h$. In other words, it is not possible to execute $\pi$ to achieve the goal with respect to $KB_h$. For example, an action in the plan cannot be executed because its precondition is not satisfied, an action in the plan does not exist, or the goal is not reached after the last action in the plan is executed. From the perspective of logic, a plan is valid if there exists at least one model in $KB_h$ in which the plan can be executed and the goal is reached:
 
 \begin{definition}
 [Plan Validity]  
 \label{val}
Given a planning problem $\Pi$, a plan $\pi$ of $\Pi$, where $\alpha_t$ is the action of the plan at time step $t$, and a knowledge base $KB_h$ encoding $\Pi$, $\pi$ is a \emph{valid} plan in $KB_h$ if $KB_h \credulous \pi \land g_n$, where $g_n$ is the fact corresponding to the goal of the planning problem at time step $n$.
 \end{definition}

\smallskip \noindent \textbf{Plan Optimality:}
Assume that $\pi^*$ is an optimal plan in a model of $KB_a$. To explain the optimality of $\pi^*$ to $KB_h$, we need to prove that no shorter (optimal) plan exists in $KB_h$. Thus, we need to prove that no shorter plan exists in \emph{all} models of $KB_h$. This can be easily done by using the notion of skeptical entailment.

\begin{definition}
[Plan Optimality]
\label{opt} 
Given a planning problem  $\Pi$, a plan $\pi$ of $\Pi$ with length $n$, and a knowledge base $KB_h$ encoding $\Pi$, the plan $\pi$ is \emph{optimal} in $KB_h$ if and only if $KB_h \credulous \pi \land g_n$ \emph{and} $KB_h \skeptical \phi$, where $\phi= \bigwedge_{t=0}^{n-1} \neg g_t$ and $g_i$ is the fact corresponding to the goal of the planning problem at time step $i$.
\end{definition}   
In essence, the query $\phi$ in the above definition is that no plan of lengths 1 to $n - 1$ exists. Therefore, when combined with the fact that a plan $\pi$ of length $n$ that achieves the goal state exists, then that plan must be an optimal plan. 

\smallskip \noindent \textbf{A Simple Approach for Restoring Consistency:}
When updating $KB_{h}$ using Definition~\ref{def:m-exp}, it might be necessary to retract some formulae from $KB_{h}$ to guarantee consistency. To efficiently solve this problem, we exploit a simple observation: There exists only a single model in a knowledge base encoding a planning problem that is consistent with an optimal plan $\pi^{*}$ for that problem. The reason is that all facts are initialized by the start state and cannot change between subsequent time steps unless there are effects of actions that are executed. Further, only one action can be taken at each time step and all actions are deterministic.
Using this observation, we generalize that the formulae in $KB_{h}$ that are false according to this model must be erroneous with respect to $KB_{a}$. A trivial approach would be to use that model to identify the erroneous formulae and replace them with the corresponding (correct) formulae from $KB_{a}$. Thus, specifying a cost-function that minimizes the complexity of an explanation, i.e., w.r.t. subset-cardinality, this framework can be used to model the MRP and yield minimal explanations. 

\smallskip \noindent \textbf{Key Differences with the Previous Framework:} The key differences with the preliminary framework proposed by \citet{vas} are the following: (1) The previous framework was restricted to skeptical entailment while our generalized version applies to credulous entailment as well. (2) The previous framework assumes that $KB_h \subseteq KB_a$, which negates the need to remove any formula from $KB_h$ during the update process to maintain consistency. In contrast, our framework makes no such assumption.

\section{Relationship to Other KR Work}
As our proposed framework bears some similarities with the theory of belief change and abductive explanations, in this section, we first describe their underlying theory. We then provide in the next section two examples that illustrate the differences between these approaches.

\subsection{Abductive Explanations}
Explanations in knowledge base systems were first introduced by \citet{levesque1989knowledge} in terms of abductive reasoning, that is, given a knowledge base and a formula that we do not believe at all, what would it take for us to believe that formula?
A more formal definition is provided below.
 
 \begin{definition}
 [Abductive Explanation]
 Given a knowledge base $KB$ and a query $q$ to be explained, $\alpha$ is an explanation of $q$ w.r.t. to $KB$ iff $KB \cup \{\alpha\}$ is consistent and $KB \cup \{\alpha\} \skeptical q$.
 \end{definition} 
Usually, such explanations are phrased in terms of a hypothesis set $H$ (set of atomic sentences -- also called abducibles), and, generally, is an intuitive methodology for deriving root causes.

\subsection{Belief Change}
Belief change is a kind of change that can occur in a knowledge base. Depending on how beliefs are represented and what kinds of inputs are accepted, different typologies of belief changes are possible. In the most common case, when beliefs are represented by logical formulae, one can distinguish three main kinds of belief changes, namely, \emph{expansion}, \emph{revision}, and \emph{contraction}. In the following, we formally describe the aforementioned notions.

\medskip \noindent \textbf{Expansion:}
An expansion operator of a knowledge base can be formulated in a logical and set-theoretic notation:

\begin{definition}
[Expansion Operator]
Given a knowledge base $KB$ and a formula $\phi$, $+_{e}$ is an expansion operator if it expands $KB$ by $\phi$ as $KB+_{e}\phi := \{ \psi : KB \cup \phi \vdash \psi \}$.
\end{definition}
It is trivial to see that $KB+_{e}\phi$ will be consistent when $\phi$ is consistent with $KB$, and that $KB+_{e}\phi$ will be closed under logical consequences.

\medskip \noindent \textbf{Revision:}
A belief revision occurs when we want to add new information into a knowledge base in such a way that, if the new information is inconsistent with the knowledge base, then the resulting knowledge base is a new consistent knowledge base. Alchourr\'{o}n, G\"{a}rdenfors, and Makinson conducted foundational work on knowledge base revision, where they proposed a set of rationality postulates, called \emph{AGM postulates}, and argued that every revision operator must satisfy them~\cite{alchourron1985logic,gardenfors1986belief,gardenfors1995belief}. Although revision cannot be defined in a set-theoretic manner closed under logical consequences (like expansion), it can be defined:

\begin{definition}
[Revision Operator]
Given a knowledge base $KB$ and a formula $\phi$, $+_{r}$ is a revision operator if it satisfies the AGM postulates for revision and modifies $KB$ w.r.t. $\phi$ such that the resulting $KB$ is consistent.
\end{definition}

%

The underlying motivation behind the AGM postulates is that when we change our beliefs, we want to retain as much as possible the information from the old beliefs. Thus, when incorporating new information in the knowledge base, the heuristic criterion should be the criterion of \emph{information economy} (i.e.,~minimal changes to the knowledge base is preferred). As such, a model-theoretic characterization of minimal change has been introduced by \citet{katsuno1991propositional}, where minimality is defined as selecting the models of $\phi$ that are ``closest'' to the models of $KB$.

However, the AGM rationality postulates will not be adequate for every application. \citet{katsuno} proposed a new type of belief revision called \emph{update}. The fundamental distinction between the two kinds of belief revision in a knowledge base, namely \emph{revision} and \emph{update}, is that the former consists of incorporating information about a static world, while the latter consists of inserting information to the knowledge base when the world described by it changes. As such, they claim that the \emph{AGM postulates describe only revision} and showed that \emph{update can be characterized by a different set of postulates called KM postulates}. 

\begin{definition}
[Update Operator]
Given a knowledge base $KB$ and a formula $\phi$, $+_{u}$ is an update operator if it satisfies the KM postulates for update and modifies $KB$ w.r.t. $\phi$ such that the updated $KB$ incorporates the change in the world introduced by $\phi$.
\end{definition}

From a model-theoretic view, the difference between revision and update, although marginal at first glance, can be described as follows: Procedures for revising $KB$ by $\phi$ are those that satisfy the AGM postulates and select the models of $\phi$ that are ``closest'' to the models of $KB$. In contrast, update methods are exactly those that satisfy the KM postulates and select, for each model $I$ of $KB$, the set of models of $\phi$ that are closest to $I$. Then, the updated $KB$ will be characterized by the union of these models.\footnote{This approach is called the \emph{possible models approach (pma)~\cite{winslett1988reasoning}.}}

To better illustrate the distinction between revision and update, let us consider the following example.

\begin{example}
Consider proposition logic theories over the set of propositional letters $\{a, b\}$ with the usual definition of models, 
satisfaction, etc. Assume a knowledge base $KB = \{ (a \wedge \lnot b)\vee (\lnot a \wedge b) \}$ and $\phi = \{a\}$. If we choose to update $KB$ with $\phi$ then, according to Katsuno's postulates, the resulting $KB$ will be $KB+_{u}\phi = \{a\}$. Now, if we choose to revise $KB$ then $KB+_{r}\phi = \{a \wedge \lnot b \}$, according to the second AGM postulate.

\end{example}

It is worth mentioning that, on a high level, the key difference between update and revision is a temporal one: Update incorporates into the knowledge base the fact that the world described by it has changed, while revision is a change to our world description of a world that has not itself changed. We refer the reader to \citet{katsuno} for a comprehensive description as well as an intuitive meaning between revision and update.

\medskip \noindent \textbf{Contraction:}
Similarly to the AGM postulates for revision, \citet{alchourron1985logic} proposed a set of axioms that any contraction operator must satisfy. Therefore, a contraction operator is defined by:

\begin{definition}
[Contraction Operator]
Given a knowledge base $KB$ and a formula $\phi$, $-_{c}$ is a contraction operator if it satisfies the AGM postulates for contraction, and contracts $KB$ w.r.t. $\phi$ by retracting formulae in $KB$ without adding of new ones. 
\end{definition}

%

Interestingly, it has been shown that the problems of revision and contraction are closely related~\cite{gardenfors1988knowledge}. Despite the fact that the postulates that characterize revision and contraction are ``independent,''\footnote{In the sense that the postulates for revision do not refer to contraction and vice versa.} revision can be defined in terms of contraction (and vice versa). This is referred to as the Levi identity~\cite{levi1978subjunctives}. 



\section{Two Illustrative Examples}

To illustrate the differences between our approach and the KR approaches described in the previous section, we discuss below how they operate on two planning examples. 

\subsection{Problem 1}

Assume a planning problem with only one action $A=\{$precondition: $P$, effect: $E\}$ with initial and goal states $P$ and $E$, respectively. Clearly, the plan for this problem is $\pi^{*} = [A]$. Also, assume that the human user is not aware that action $A$ has effect $E$. Now, the knowledge bases encoding the models of the agent and the human, in the fashion of~\citet{kautz1996encoding}, are:
\beitemize
\item $KB_a = [P_0, \lnot E_0, E_1, A_0 \rightarrow P_0, A_{0} \rightarrow E_{1}, \lnot E_{0} \wedge E_{1} \rightarrow A_{0}],$
\item $KB_h = [P_0, \lnot E_0, E_1, A_0 \rightarrow P_0].$
\enitemize
Further, without loss of generality, suppose that the explanation needed to explain $\pi^{*}$ to $KB_{h}$ is $\epsilon = [A_{0} \rightarrow E_{1}, \lnot E_{0} \wedge E_{1} \rightarrow A_{0}]$.

\medskip \noindent \textbf{Abductive Explanations:}
Abductive explanation cannot be applied in this setting because $KB_{h}$ does not contain any causal rules that can be used to abduce the query.

\medskip \noindent \textbf{Revision:}
Since the union of $\epsilon$ and $KB_{h}$ is consistent, the revision operator will yield a trivial update according to the second AGM axiom: 
$KB_h +_r \epsilon = KB_h \cup \epsilon$. 

\medskip \noindent \textbf{Update:}
To use the update operator, we first need to find the models of $KB_{h}$ and $\epsilon$:
\beitemize
\item[] \emph{Models}($KB_h$):   $I_1 = \{P_0, E_1, A_0\}$, $I_2 = \{P_0, E_1\}$.

\item[] \emph{Models}($\epsilon$):  $J_1 = \{A_0, E_1, P_0\}, J_2 = \{A_0, E_1\},$ \\ 
$J_3 = \{A_0, E_1, E_0, P_0\}, J_4 = \{A_0, E_1, E_0 \}$, \\
$J_5 = \{E_1, E_0, P_0\}, J_6 = \{E_1, E_0\}$, \\
$J_7 = \{E_0, P_0 \}, J_8 = \{E_0 \}, $
$J_9 = \{P_0 \},  J_{10} = \{ \}$. 
\enitemize
Now, according to the KM postulates, we need to find the models of $\epsilon$ that are closest to $I_{1}$ and $I_{2}$. Then, the updated $KB$ will be the disjunction of the conjunction of the variables in each model. Now, let the function $\Diff(m_1, m_2)$ denote the set of propositional letters with different truth values in models $m_1$ and $m_2$.

Then, for $I_1$, it is easy to see that the closest model is $J_{1}$ because $\Diff(I_1,J_1) = \emptyset <\Diff(I_1,J_k)$ for all $k$. So, $J_{1}$ is selected. For $I_{2}$, we need to calculate the difference for every model of $\epsilon$:
\beitemize
\item[]
\underline{$\Diff(I_2,J_1) = \{A_0\}$}, \hspace{1em} $\Diff(I_2,J_2) = \{A_0, P_0\}$, \\
$\Diff(I_2,J_3) = \{A_0,  E_0\}$, 	\hspace{1em}$\Diff(I_2,J_4) = \{A_0, E_0, P_0\}$, \\
\underline{$\Diff(I_2,J_5) = \{ E_0\}$}, \hspace{1em} $\Diff(I_2,J_6) = \{P_0, E_0\}$, \\
$\Diff(I_2,J_7) = \{E_0, E_1 \}$, \hspace{0.5em} $\Diff(I_2,J_8) = \{E_0,E_1,P_0 \}$, \\
\underline{$\Diff(I_2,J_9) = \{E_1\}$}, \hspace{1em} $\Diff(I_2,J_{10}) = \{P_0,E_1\}$,
\enitemize
where sets with the minimal elements are underlined. So, $J_1$, $J_5$, and $J_9$ are selected and the final result is the union of all selected models, that is, \emph{Models}($KB_h +_{u} \epsilon$) $= \{J_1, J_5, J_9\}$. Thus, the resulting $KB$ must satisfy all three models, yielding the following: $KB_h +_{u} \epsilon = [\big(A_{0} \wedge E_{1} \wedge P_{0} \wedge \lnot E_{0} \big) \vee \big(E_{1} \wedge E_{0} \wedge P_{0} \wedge \lnot A_{0} \big) \vee \big(P_{0} \wedge \lnot E_{1} \wedge \lnot A_{0} \wedge \lnot E_{0} \big)]$.

\medskip \noindent \textbf{Our Approach:}
As a first step, our method will first check if $KB_{h}$ is consistent with the model of $KB_{a}$. Since it is, it will simply insert $\epsilon$ to $KB_{h}$, yielding $\widehat{KB}_h^{\epsilon} = KB_h \cup \epsilon$ just like revision.

In conclusion, this problem demonstrates that it is possible for \emph{belief revision} to yield the same update as our approach, which is when $KB_h \cup \epsilon$ is consistent (per AGM postulates). It also highlights why \emph{belief update} is not applicable for explainable planning, namely that the updated knowledge base $KB_{h} +_{u} \epsilon$ violates the action dynamics of planning problems \cite{kautz1996encoding}.

\subsection{Problem 2}
Now assume a planning problem with the two actions $A=\{$precondition: $P$, effect: $G\}$ and $B=\{$precondition: $E$, effect: $G\}$ with initial and goal states $P$ and $G$, respectively, and a plan $\pi^{*} = [A]$. Also, assume that the human user is not aware that action $A$ has effect $G$. Then, the knowledge bases encoding the models of the agent and the human are:
\beitemize
\item $KB_a =[P_0, \lnot E_0, \lnot G_0, G_1, A_0 \rightarrow P_0, A_0 \rightarrow G_1, B_0 \rightarrow E_0, B_0 \rightarrow G_1, \lnot G_0 \wedge G_1 \rightarrow A_0 \vee B_0, \lnot A_0 \vee \lnot B_0],$
\item $KB_h = [P_0, \lnot E_0, \lnot G_0, G_1, A_0 \rightarrow P_0, B_0 \rightarrow E_0, B_0 \rightarrow G_1, \lnot G_0 \wedge G_1 \rightarrow B_0, \lnot A_0 \vee \lnot B_0].$
\enitemize
As in the previous problem, we now assume that the explanation needed is $\epsilon = [A_0 \rightarrow G_1, \lnot G_0 \wedge G_1 \rightarrow A_0 \vee B_0].$

\medskip \noindent \textbf{Abductive Explanations:}
The method of abductive explanations will fail in this setting due to the fact that $KB_{h}$ is inconsistent. Further, even if it is consistent, we will still not be able to find any abductive explanations due to the lack of causal rules in $KB_{h}$.

\medskip \noindent \textbf{Revision:}
Following AGM postulates, revision cannot be applied because $KB_{h}$ is individually inconsistent.

\medskip \noindent \textbf{Update:}
Again, as $KB_h$ is inconsistent, and according to KM update postulates, it cannot be repaired using update.

\medskip \noindent \textbf{Our Approach:} As $KB_h \cup \epsilon$ is inconsistent, our approach will identify the erroneous formula $\lnot G_0 \wedge G_1 \rightarrow B_0$ and replace it with the corresponding correct formula $\lnot G_0 \wedge G_1 \rightarrow A_0 \vee B_0$ from $KB_a$, thereby restoring consistency. The updated knowledge base will be $\widehat{KB}_h^{\epsilon} = [P_0, \lnot E_0, \lnot G_0, G_1, A_0 \rightarrow P_0, A_0 \rightarrow G_1, B_0 \rightarrow E_0, B_0 \rightarrow G_1, \lnot G_0 \wedge G_1 \rightarrow A_0 \vee B_0, \lnot A_0 \vee \lnot B_0]$.

In summary, this problem demonstrates that when $KB_h$ is inconsistent, abductive explanations, revision, and update cannot be applied but our approach can be applied.

\section{Discussion and Conclusions} 

A key distinction between the previous approaches and our approach is that, historically, belief change refers to a \emph{single agent} revising its belief after receiving a new piece of information that is in conflict with its current beliefs; so, there is a temporal dimension in belief change and a requirement that it should maintain as much as possible the belief of the agent, per AGM postulates. Our notion of explanation is done with respect to \emph{two knowledge bases} and there is no such requirement (with respect to $KB_h$). For example, if the agent believes that block A is on block B, the human believes that block B is on block A, and the explanation does not remove this fact from the human's KB, then the agent and the human will still have some conflicting knowledge about the positions of blocks A and B after the update. Thus, the previous notions of belief change cannot accurately capture and characterize the MRP problem. 

Similar to belief change, explanation differs from other similar notions, such as diagnosis~\citep{rei87}. In general, a diagnosis is defined with respect to a knowledge base $KB$, a set of components $H$, and a set of observations $O$. Given that $KB \cup O \cup \{\neg ab(c) \mid c \in H\}$ is inconsistent, a diagnosis is a subset $S$ of $H$ such that $KB \cup O \cup \{ab(c) \mid c \in S\} \cup \{\neg ab(c) \mid c \in H\setminus S\}$ is consistent. Here, $ab(c)$ denotes that the component $c$ is faulty. 

Generalizing this view, the inconsistency condition could be interpreted as the query $q$ and $KB \cup O \skeptical \neg q$. Then a diagnosis is a set $S \subseteq H$ such that $KB \cup O \cup S \skeptical q$. An explanation for $q$ from $KB_a$ to $KB_h$ is, on the other hand, a pair $(S_1, S_2)$ such that $(KB_h \setminus S_2) \cup S_1 \skeptical q$. Thus, the key difference is that an explanation might require the removal of some knowledge of $KB_h$ while a diagnosis does not.

A recent research direction that is closely related to the proposed notion of explanation is that by \citet{shvo-etal-extraamas20}, where they propose a general belief-based framework for generating explanations that employs epistemic state theory to capture the models of the explainer (agent in this paper) and the explainee (human user in this paper), and incorporates a belief revision operator to assimilate explanations into the explainee's epistemic states. A main difference with our proposed framework is that our framework restricts knowledge to be stored in logical formulae, while theirs considers epistemic states that can characterize different types of problems and have no such restriction. 

To conclude, we build upon notions from KR to expand a preliminary logic-based framework that characterizes the model reconciliation problem for explainable planning. We further provide a detailed exposition on the relationship between our framework and other similar KR techniques, such as abductive explanations and belief change, using two illustrative planning examples to describe their differences. 

As our explanations are currently represented in logical formulae, specifically in propositional logic, in our future work, we aim to translate them into a human-understandable format (e.g.,~by employing natural language processing techniques). Further, we plan to generalize our framework to account for problems beyond classical planning problems, such us hybrid planning (PDDL+ \cite{fox2006modelling}), as long as these problems can be encoded using a logical formalism.



\section*{Acknowledgment}

This research is partially supported by NSF grants 1345232, 1757207, 1812619, 1812628, 1914635, and NIST grant 70NANB19H102. The views and conclusions contained in this document are those of the authors and should not be interpreted as representing the official policies, either expressed or implied, of the sponsoring organizations, agencies, or the U.S. government.

\bibliographystyle{named}  
\bibliography{ref}

\end{document}